\pdfoutput=1
\documentclass[11pt]{article}

\usepackage{acl}

\usepackage{times}
\usepackage{latexsym}

\usepackage[T1]{fontenc}
\usepackage[utf8]{inputenc}

\usepackage{microtype}

\usepackage[main=english, czech, german, latin]{babel}
\usepackage{tikz}
\usetikzlibrary{positioning}
\usepackage[citations,
            contentBlocks,
            relativeReferences,
            taskLists]{markdown}
\usepackage{booktabs}
\usepackage[all]{nowidow}
\usepackage{paralist}
\usepackage{subfig}
\usepackage{tabularx}
\usepackage{ragged2e}

\newcommand{\Adaptor}{Adapt\scalebox{0.7}{$\mathcal{O}$}r}
\makeatletter
\ifacl@finalcopy
  \newcommand{\AHISTO}{AHISTO}
\else
  \newcommand{\AHISTO}{HUSSITE}
\fi
\makeatother

\title{People and Places of Historical Europe: 
Bootstrapping Annotation Pipeline and a New Corpus of Named Entities in Late Medieval Texts}

\newcommand\inst[1]{\textsuperscript{\textnormal{#1}}}
\author{%
  \vspace{5pt}Vít Novotný\inst{1} \and Kristýna Luger\inst{2} \and Michal Štefánik\inst{1} \\[-4pt] \textbf{Tereza Vrabcová}\inst{1} \and \textbf{Aleš Horák}\inst{1} \\[10pt]
  \inst{1}Faculty of Informatics, Masaryk University, Brno, Czech Republic \\
  \inst{2}Faculty of Arts, Masaryk University, Brno, Czech Republic \\
  \texttt{\{witiko,449852,stefanik.m,485431,hales\}@mail.muni.cz} \\
}

\begin{document}

\maketitle

\begin{abstract}
% Background
Although pre-trained named entity recognition (NER) models are highly accurate on modern corpora, they underperform on historical texts due to differences in language OCR errors.
% Aims
In this work, we develop a new NER corpus of 3.6M sentences from late medieval charters written mainly in Czech, Latin, and German.

% Results
We show that we can start with a list of known historical figures and locations and an unannotated corpus of historical texts, and use information retrieval techniques to automatically bootstrap a NER-annotated corpus. Using our corpus, we train a NER model that achieves entity-level Precision of 72.81--93.98\% with 58.14--81.77\% Recall on a manually-annotated test dataset. Furthermore, we show that using a weighted loss function helps to combat class imbalance in token classification tasks. To make it easy for others to reproduce and build upon our work, we publicly release our corpus, models, and experimental code.
\end{abstract}

\section{Introduction}
\label{sec:introduction}

Named entity recognition (NER) techniques enable the extraction of valuable insights from unstructured information in various domains. With the advancements in optical character recognition (OCR, ~\citealp{breuel2017high,kodym2021page}), NER can now be applied to scanned historical texts spanning over a millennium. However, despite the significant interest in NER, resources for training models to recognize entities in medieval texts remain scarce~\cite[Table~3]{ehrmann2021named}.

In this work, we present a new multilingual NER corpus of 3.6M sentences from the \AHISTO{} project\makeatletter\ifacl@finalcopy\else\footnote{The project name was changed to maintain anonymity.}\fi\makeatother, which aims to build a searchable web database of late medieval charters from scanned images. In Section~\ref{sec:related-work}, we review recent related work in historical NER. In Section~\ref{sec:data-description} and \ref{sec:corpus-annotation}, we describe the database and the automatic pipeline used to bootstrap our corpus in. In Section~\ref{sec:experiments}, we use our corpus to train models for historical NER and evaluate them on a manually-annotated test dataset. We show that our models are highly-accurate, reaching Precision of up to 94\% with up to 82\% recall. Additionally, we show that the use of a weighted loss function is crucial in token classification tasks with high class imbalance. We conclude in Section~\ref{sec:conclusion}.

To facilitate reproducibility and future research, we publicly release our corpus under open CC0 license as well as our models and code.\footnote{\url{https://nlp.fi.muni.cz/projects/ahisto/ner-resources}}

\section{Related Work}
\label{sec:related-work}

\citet{grover-etal-2008-named} created a corpus of British parliamentary proceedings from the late 17th and early 19th centuries using OCR techniques and expert annotation. They developed and evaluated a rule-based NER classifier, achieving an F$_1$-score of 71\%, noting the detrimental effect of OCR errors.

\citet{hubkova2020czech} created a corpus of 32 historical Czech newspapers from 1872 using OCR and expert annotation. \citet{hubkova2021transfer} achieved a state-of-the-art F$_1$-score of 82\% on the corpus by fine-tuning a pre-trained SlavicBERT model~\cite{arkhipov-etal-2019-tuning}.

\citet{blouin-etal-2021-transferring} fine-tuned pre-trained monolingual Transformer models on early modern NER corpora in English, French, and German, achieving a near-state-of-the-art F$_1$-score of 62\%. They also found that the character-based CharBERT model~\cite{ma-etal-2020-charbert} was more robust against OCR errors than BERT~\cite{devlin-etal-2019-bert}.

\citet{torres-aguilar-2022-multilingual} developed a corpus of 7,576 medieval charters ranging from the 10th century to the 15th century using expert annotation. They showed that finetuning pre-trained multilingual Transformer models such as mBERT~\cite{devlin-etal-2019-bert} and XLM-RoBERTa~\cite{conneau2020unsupervised} gave comparable results to state-of-the-art Bi-LSTM-CRF NER models~\cite{ma-hovy-2016-end}.

\section{Data Description}
\label{sec:data-description}

\begin{figure}
\shorthandoff{-}
\centering
\scalebox{0.73}{\begin{tikzpicture}

\node (deed) [draw] at (0, 0) { Charter };
\node (book) [draw, right = 4cm of deed] { Book };
\node (regest) [draw, below = 4cm of deed] { Abstract };
\node (entity) [draw, below = 4cm of book] { Entity };

\draw (deed.east) node [xshift =  12pt, yshift =  7pt] { 0..* } to
      (book.west) node [xshift = -12pt, yshift = -7pt] { 0..* }
                  node [xshift = -31pt, yshift =  7pt] { $\triangleleft$ \emph{transcribes} } ;

\draw (deed.south)   node [xshift = -6pt, yshift = -12pt] { 1 } to
      (regest.north) node [xshift =  6pt, yshift =  12pt] { 1 }
                     node [xshift = -7pt, yshift =  29pt] { \rotatebox{90}{\emph{describes} $\triangleright$} } ;

\draw (book.south)   node [xshift =  6pt, yshift = -12pt] { 1 } to
      (entity.north) node [xshift = 10pt, yshift =  12pt] { 0..* }
                     node [xshift = -7pt, yshift =  29pt] { \rotatebox{90}{\emph{occurs in} $\triangleright$} } ;

\draw (regest.east) node [xshift =  12pt, yshift = -7pt] { 0..* } to
      (entity.west) node [xshift = -12pt, yshift =  7pt] { 0..* }
                    node [xshift = -26pt, yshift = -7pt] { \emph{$\triangleleft$ occurs in} } ;

\draw (regest.north east) node [xshift =  20pt, yshift =  28pt] { \rotatebox{45}{0..*} }
                          node [xshift =  32pt, yshift =  20pt] { \rotatebox{45}{\emph{cites} $\triangleright$} } to
      (book.south)        node [xshift = -20pt, yshift = -28pt] { \rotatebox{45}{0..*} } ;

\node (book-text) [draw, right = 1cm of book, circle, yshift = 6pt] { text } ;
\node (entity-type) [draw, right = 0.9cm of entity, circle, yshift = -4pt] { type } ;
\node (regest-text) [draw, left = 1cm of regest, circle, yshift = -5pt] { text } ;

\draw (book) to (book-text) ;
\draw (entity) to (entity-type) ;
\draw (regest) to (regest-text) ;

\end{tikzpicture}}
\caption{Entity relationship diagram of the database that was produced in the \AHISTO{} project.}
\label{fig:erd}
\vspace{-0.335cm}
\end{figure}

The \AHISTO{} project database includes various document types, as shown in Figure~\ref{fig:erd}. The main focus is on \emph{charters} from Europe during the Hussite era (1419–1436), each with an accompanying \emph{abstract} written by project historians in contemporary Czech. Historians also manually annotate all named entities (people and places) in the abstracts. The original text of charters, if available, can be found in \emph{books}. OCR texts of books are available\makeatletter\ifacl@finalcopy~\cite{novotny2021when,novotny2022when}\fi\makeatother, but named entities are not annotated.

\looseness=-1%  Make the paragraph one line shorter.
The database includes 4,182 abstracts with 5,621 sentences, and 15,100 unique named entities: people (62.53\%) and places (37.47\%). It also includes 872 books with 268,669 pages and 3.6M sentences, mostly in medieval Czech (43.23\%), Latin (36.32\%), and German (16.89\%). Of these pages, 3,553 (1.32\%) with 50k sentences were selected as relevant to medieval charters by project historians.

\section{Corpus Annotation}
\label{sec:corpus-annotation}

We developed five NER corpora from the database of the \AHISTO{} project. We created \textbf{Abstracts-Tiny} from abstracts and \textbf{Books-Small}, \textbf{Medium}, \textbf{Large}, and \textbf{Huge} from books. The statistics of all our corpora are listed in Table~\ref{tab:corpus-statistics}.

\begin{table}
\caption{The numbers of sentences and the occurrences of people (B-PER tokens) and places (B-LOC tokens) in our NER corpora. For each corpus, we report statistics for training, validation, and testing splits.}
\label{tab:corpus-statistics}
\centering
\small
\begin{tabular}{l@{}rrrr}
\toprule
Corpus & \# Sentences & \# B-PER & \# B-LOC \\
\midrule
\textbf{Abstracts-Tiny} & \bf 5,222 & \bf 10,981 & \bf 5,933 \\
 \quad Training   &     4,320 &     9,032 &     4,952 \\
 \quad Validation &       502 &     1,160 &       606 \\
 \quad Testing    &       400 &       789 &       375 \\
\textbf{Books-Small} & \bf 7,842 & \bf 4,778 & \bf 5,679 \\
 \quad Training   &     6,493 &     3,877 &     4,722 \\
 \quad Validation &     1,249 &       614 &       714 \\
 \quad Testing    &       100 &       287 &       243 \\
\textbf{Books-Medium} & \bf 7,842 & \bf 17,400 & \bf 17,184 \\
 \quad Training   &     6,493 &    13,958 &    13,987 \\
 \quad Validation &     1,249 &     3,155 &     2,954 \\
 \quad Testing    &       100 &       287 &       243 \\
\textbf{Books-Large} & \bf 46,739 & \bf 46,051 & \bf 45,435 \\
 \quad Training   &    44,155 &    43,360 &    42,315 \\
 \quad Validation &     2,484 &     2,404 &     2,877 \\
 \quad Testing    &       100 &       287 &       243 \\
\textbf{Books-Huge} & \bf 3,629,903 & \bf 4,257,380 & \bf 2,865,470 \\
 \quad Training   & 3,227,624 & 3,794,991 & 2,545,820 \\
 \quad Validation &   402,179 &   462,102 &   319,407 \\
 \quad Testing    &       100 &       287 &       243 \\
\bottomrule
\end{tabular}
\vspace{-0.335cm}
\end{table}

\subsection{Corpus from Abstracts}
\label{sec:corpus-from-abstracts}

For the evaluation of NER models on contemporary Czech texts that discuss medieval charters, we constructed a corpus \textbf{Abstracts-Tiny} from all 5,621 sentences in abstracts. We randomly split the sentences into 80\%, 10\%, and 10\% for training, validation, and testing.

\subsection{Bootstrapping Initial Corpus from Books}
\label{sec:bootstrapping-initial-corpus-from-books}

To bootstrap our initial corpus \textbf{Books-Small}, we used the information retrieval system from the Manatee library~\cite{rychly2007manatee,busta2023nosketchengine}, which performed the best out of 9 systems that we considered, see also Appendix~\ref{app:comparison-of-bootstrapping-methods}. First, we indexed OCR texts from the 3,553 book pages that historians selected as relevant. Then, for each of the 15,100 named entities in abstracts, we used a boolean phrase query to retrieve all occurrences of the named entity in the index. For each occurrence of a named entity, we extracted the surrounding sentence and we merged all sentences that were extracted multiple times.

We randomly split the sentences into 80\%, 10\%, and 10\% for training, validation, and testing of NER models. From the testing split, we randomly sampled 100 sentences and a volunteer Czech graduate student of history manually checked all entities in the sentences. See Appendix~\ref{app:annotator-instructions} for annotator instructions. We used the 100 sentences for testing.

\subsection{Inferring Missing Entities in Books}
\label{sec:inferring-missing-entities-in-books}

Sentences in the \textbf{Books-Small} corpus contained many named entities that were not part of abstracts and were therefore missing. To produce our intermediate \textbf{Books-Medium} corpus, we trained a NER model on the \textbf{Books-Small} corpus and used it to infer the missing named entities in \textbf{Books-Small}.

Most named entities in the \textbf{Books-Medium} corpus were annotated, but the corpus only contained a small portion of the 3.6M sentences in all books. To produce our final \textbf{Books-Large} corpus, we trained a NER model on the \textbf{Books-Medium} corpus and used it to infer named entities in all book pages that historians selected as relevant. Furthermore, we also inferred named entities in all books to produce the corpus \textbf{Books-Huge}, which is $100\times$ larger than \textbf{Books-Large} but may contain irrelevant sentences.

We describe the NER models that we used for the inference in the following section.

\section{Experiments}
\label{sec:experiments}

In this section, we describe the NER models that we used to produce the \textbf{Books-Medium}, \textbf{Large}, and \textbf{Huge} corpora and how we evaluated them.

\subsection{Models}
\label{sec:models}

We trained two models by fine-tuning pretrained XLM-RoBERTa models~\cite{conneau2020unsupervised} using the \Adaptor{} library~\cite{stefanik2022adaptor} for multi-objective training.

To produce the \textbf{Books-Medium} corpus, we fine-tuned the XLM-RoBERTa-Base model (125M parameters). To produce the \textbf{Books-Large} and \textbf{Books-Huge} corpora, we fine-tuned the XLM-RoBERTa-Large model (355M parameters). To simplify the discussion, we will refer to the models as \textbf{Model S} (for small) and \textbf{Model L} (for large) throughout the paper. See Appendix~\ref{app:training-details} for the description of our hardware and hyperparameters.

Given the scarcity of pre-trained historical NER models that are publicly available, we compare our model against the XLM-RoBERTa-Large model fine-tuned on contemporary German news texts~\cite{gugger2022xlm} from the CoNLL03 dataset~\cite{tjong-kim-sang-de-meulder-2003-introduction}. Although the model was only trained on German data, \citet{ruder-etal-2019-unsupervised} show that the model should generalize well to other languages.

\subsection{Objectives}
\label{sec:objectives}

In order to train our models effectively, we used a multi-objective approach, utilizing two distinct objectives in our optimization process:
\begin{compactdesc}
\item[Masked Language Modeling (MLM):] Unsuper\-\textls[-7]{vised regression on the \textbf{Books-Large} corpus.}
\item[Token Classification (TC):] Supervised classification of tokens into classes B-PER, I-PER, B-LOC, I-LOC, and O. To address the issue of class imbalance, we use the weighted cross-entropy (WCE) loss function with inverse class frequencies as weights.
\end{compactdesc}

We adopt a sequential schedule for alternating between objectives, where each objective is trained for a single epoch. This approach has the advantage of allowing focused optimization of each objective.

\subsection{Quantitative Evaluation}
\label{sec:quantitative-evaluation}

We use the micro-averaged token-level F$_\beta$-score with a value of $\beta=0.25$ as the evaluation metric for both the validation of the TC objective and the quantitative evaluation of our models. This metric prioritizes precision in entity recognition, even if it results in lower recall. To evaluate out-of-domain performance on contemporary Czech texts that discuss medieval charters, we report the F$_\beta$-score on two benchmarks: the \textbf{Abstracts-Tiny} corpus and the \textbf{Books-*} corpora.

In addition to the token-level $F_\beta$-score, we also present the entity-level Precision and Recall. These measures are considered to be more representative for the majority of NER applications, as opposed to per-token evaluation measures, which are tied to the tokenizer of a model. Similarly to \citet{ehrmann2020overview}, we use two evaluation regimes:
\begin{compactdesc}
\item[Strict] Predicted entities must match both the type and boundaries of expected entities.
\item[Fuzzy] Predicted entities must match the type and overlap the boundaries of any expected entity.
\end{compactdesc}
We report the micro-averaged Precision and Recall for our best model as a range of Strict--Fuzzy\%. We report both overall and per-language results.

\subsection{Qualitative Evaluation}
\label{sec:qualitative-evaluation}

In order to conduct a comprehensive qualitative evaluation, we also report a confusion matrix of our best model on the \textbf{Books-*} corpora. In Appendix~\ref{app:model-predictions}, we also compare the predictions made by our best model with manual annotations.

\subsection{Ablation Study}
\label{sec:ablation-study}

In this section, we present a series of ablation experiments designed to investigate the impact of using various training data and loss functions for the TC objective. In the experiments, we fine-tune the small XLM-RoBERTa-Base model due to environmental considerations.

\paragraph{Training Data Size}
We evaluate the validity of our annotations by training the TC objective not only on the \textbf{Books-Medium} corpus, but also the \textbf{Books-Small} and \textbf{Large} corpora. We will refer to the models as \textbf{Model~TDS1} and \textbf{Model~TDS2}.

\paragraph{Class Imbalance}
We replaced the weighted cross-entropy loss function in the TC objective with unweighted cross-entropy loss in the \textbf{Books-Small} corpus to investigate the impact of missing named entities. We will refer to this model as \textbf{Model~CI}.

\section{Discussion}
\label{sec:discussion}

\subsection{Quantitative Evaluation}

\begin{table}
\caption{Evaluation results for our NER models and the baseline. For each model, we list a short identifier for ease of reference, the size of the model in parameters, the training data and the loss function of the Token Classification (TC) objective, the loss function used in the TC objective, and the accuracy measured by per-token F$_\beta$-score on both the \textbf{Abstracts-Tiny} corpus and the \textbf{Books-*} corpora. Best results are \textbf{bold}.}
\label{tab:quantitative-results}
\centering
\small
\setlength\tabcolsep{3pt}
\begin{tabular}{@{}c@{\kern 2pt}c@{\kern 2.5pt}ccc@{\kern4pt}c@{}}
\toprule
Model & Training & Model & TC & \multicolumn{2}{c}{F$_\beta$-score} \\
Id & Data & Size & Loss & \textls[-25]{Abstracts-Tiny} & \textls[-25]{Books-*} \\
\midrule
L    & \textls[-25]{Books-Medium} & 355M & WCE & \textbf{91.61\%} & \textbf{93.43\%} \\
S    & \textls[-25]{Books-Medium} & 125M & WCE & 90.51\% & 93.19\% \\
TDS2 & \textls[-25]{Books-Large}  & 125M & WCE & 87.21\% & 89.11\% \\
TDS1 & \textls[-25]{Books-Small}  & 125M & WCE & 88.92\% & 88.20\% \\
CI   & \textls[-25]{Books-Small}  & 125M & CE  & 86.41\% & 84.45\% \\
     & \textls[-25]{CoNLL03 (de)} & 355M & CE  & 80.59\% & 80.74\% \\
\bottomrule
\end{tabular}
\end{table}

Table~\ref{tab:quantitative-results} shows that both \textbf{Model L} and \textbf{Model S} achieved a per-token F$_\beta$-score of over 90\% on both benchmarks, outperforming the baseline model by more than 10\% on both benchmarks.

Despite its smaller size, \textbf{Model S} received only 1.1\% less per-token F$_\beta$-score than \textbf{Model L} on the \textbf{Abstracts-Tiny} corpus and only 0.24\% less on the \textbf{Books-*} corpora. This makes \textbf{Model S} a compelling choice for low-resource NER applications.

\textls[-14]{The per-entity Precision of \textbf{Model L} on the \textbf{\mbox{Books-*}} corpora was 72.81--93.98\% with 58.14--81.77\% Recall. This shows that our model can reliably recognize named entities, even though it does not always exactly match the boundaries. Per-language, Precision was the highest for Czech (77.42--95.63\%) and the lowest for German (70.80--87.07\%).}

\subsection{Qualitative Evaluation}
\label{sec:qualitative-discussion}

\begin{figure}
\centering
\includegraphics[width=\linewidth]{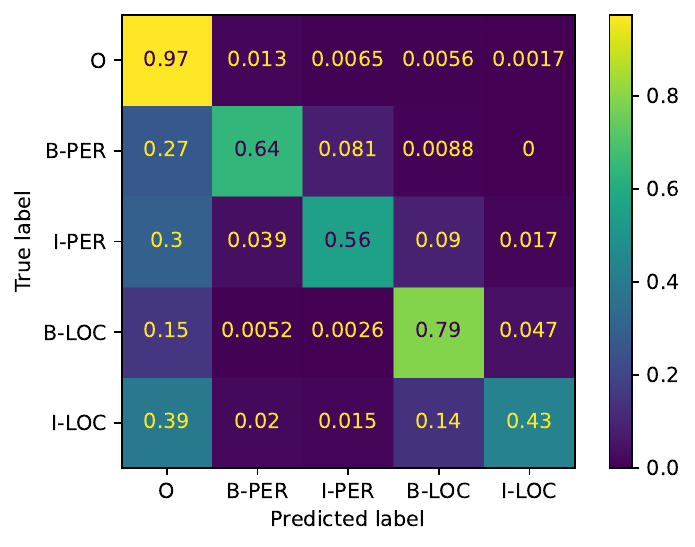}
\vspace{-0.75cm}
\caption{A confusion matrix of our model L on the \textbf{Books-*} corpora. Best viewed in color.}
\label{fig:confusion-matrix}
\end{figure}

Figure~\ref{fig:confusion-matrix} reveals that \textbf{Model L} is more accurate at identifying the beginnings of named entities than their endings. This is evidenced by the smaller probability of misclassifying tokens B-PER and B-LOC as class~O ($\leq$ 21\%) compared to the I-PER and I-LOC tokens ($\geq$ 28\%).

Our model will also sometimes recognize a list of places as a single named entity or divide a place name into several named entities. This is evidenced by the relatively high probability of classifying tokens B-LOC as I-LOC (16\%) and vice versa (20\%).

Our model will seldom make a mistake in identifying the type of a named entity. This is evidenced by the low probability of misclassifying tokens \mbox{*-PER} as \mbox{*-LOC} ($\leq$ 1\%) and vice versa ($\leq$ 2.4\%).

\subsection{Ablation Study}

Table~\ref{tab:quantitative-results} shows that both \textbf{Model TDS1} and \textbf{Model TDS2} performed worse than \textbf{Model L} on both benchmarks. \textbf{Model TDS1} was affected by the class imbalance in the \textbf{Books-Small} corpus, missing many named entities. \textbf{Model TDS2} suggests that even though the \textbf{Books-Large} corpus was constructed from pages relevant to medieval charters, there may be irrelevant sentences within the pages.

\textbf{Model CI} received 2.51\% less per-token F$_\beta$-score than \textbf{Model TDS1} on the \textbf{Abstracts-Tiny} corpus and 3.75\% less on the \textbf{Books-*} corpora. This shows that using weighted loss is crucial in token classification tasks with high class imbalance.

\section{Conclusion}
\label{sec:conclusion}

Despite the large interest in named entity recognition (NER) in the last few decades, studies targeting late medieval historical texts are still scarce.

\looseness=-1%  Make the paragraph one line shorter.
In our work, we have developed a new silver-standard NER corpus of 3.6M sentences from late medieval charters. We described our automatic pipeline for bootstrapping a corpus using a list of known named entities. We also showed that our corpus can be used to train highly accurate models for historical NER. Lastly, we have demonstrated the usefulness of using weighted loss functions in token classification tasks with high class imbalance.

\section{Limitations}
\label{sec:limitations}

In this study, we assessed the quality of the corpora \textbf{Books-Small}, \textbf{Medium}, and \textbf{Large}, by training and evaluating a NER model on them but we did not include the corpus \textbf{Books-Huge} in our analysis. However, our results on the \textbf{Books-Large} corpus indicate that there is no substantial benefit to using a corpus larger than \textbf{Books-Medium} for training NER models. This is consistent with prior research on few-shot training on smaller corpora achieving comparable accuracy to larger, potentially noisy corpora~\cite{blouin-etal-2021-transferring}.

\looseness=-1%  Make the paragraph one line shorter.
Given the scarcity of benchmarks for late medieval NER~\cite[Table~3]{ehrmann2021named}, we were unable to conduct experiments on corpora other than our own. Additionally, we utilized a NER model trained on contemporary texts as our baseline for comparison. Therefore, it is important to note that these results may not generalize to other medieval NER tasks. In the future, efforts should be made to develop more comprehensive benchmarks for late medieval NER such as our own.

\section{Ethical Considerations}

In conducting this research, we were committed to upholding the ethical principles of respect for persons, beneficence, and non-maleficence. Specifically, the annotation in this work was carried out voluntarily by informed participants, and the welfare and rights of these participants were protected throughout the research process. Furthermore, the annotation did not involve collecting any personal or sensitive information from individuals.

\makeatletter
\ifacl@finalcopy
\section*{Acknowledgements}
This work has been produced with the assistance of \href{https://sources.cms.flu.cas.cz/}{the Czech Medieval Sources online database}, provided by \href{https://lindat.cz/}{the LINDAT/CLARIAH-CZ research infrastructure}, supported by the Ministry of Education, Youth, and Sports of the Czech Republic (Project No. \href{https://starfos.tacr.cz/en/project/LM2018101}{LM2018101}).
\fi

\bibliography{anthology,custom}
\bibliographystyle{acl_natbib}

\appendix

\section{Comparison of Bootstrapping Methods}
\label{app:comparison-of-bootstrapping-methods}

For all the 15,100 named entities in abstracts, we used several information retrieval techniques to find their occurrences in the books.

First, we used four fast techniques to produce lists of up to 10,000 candidate results:

\begin{enumerate}
\item \textbf{Jaccard Similarity}~\cite{deng2015unified}: From the books, we extracted substrings of length that were similar to the length of the current entity. We ordered the substrings by their character and word Jaccard similarity to the current entity.
\item \textbf{Okapi BM25}~\cite{robertson1995okapi}: From the books, we extracted phrases of length that were similar to the length of the current entity and we indexed them as individual retrieval units in an inverted index. We retrieved the phrases using a ranked retrieval query for the current entity, using BM25 weighting.
\item \textbf{Manatee}~\cite{rychly2007manatee,busta2023nosketchengine}: From the books, we extracted lemmatized tokens and we indexed them as individual retrieval units in a positional inverted index. We retrieved phrases using a boolean phrase query for the current entity. Since boolean retrieval results are not ranked, we ranked them by their character edit distance to the current entity.
\item \textbf{Fuzzy Regexes}~\cite{navarro2003approximate}: From the current entity, we extracted an approximate regular expression. From the books, we retrieved all substrings that matched the regular expression up to a certain edit distance.
\end{enumerate}

\noindent Then, we used three slow techniques to rerank all candidate results for each named entity:

\begin{enumerate}
\item \textbf{Edit Distance}: We reordered the results by their word and character edit distance to the current entity.
\item \textbf{BERTScore}~\cite{zhang2020bertscore}: We reordered the results by their BERT F$_1$-score to the current entity.
\item \textbf{SentenceBERT}~\cite{reimers2019sentence}: We reordered the results by the cosine similarity of their SentenceBERT embeddings to the embedding of the current entity.
\end{enumerate}

\noindent Finally, we used two rank fusion techniques to combine the results of the above techniques:

\begin{enumerate}
\item \textbf{Reciprocal Rank Fusion}~\cite{cormack2009reciprocal}: We combined the results of all the inexpensive and expensive techniques based on the ranks of the results across the techniques.
\item \textbf{Concatenation}: We started with the results produced by Fuzzy Regexes, if any, followed by the results produced by the Reciprocal Rank Fusion. For duplicate results, we kept the results from Fuzzy Regexes.
\end{enumerate}

\begin{table}
\caption{The Precision, Recall, and F$_\beta$-score ($\beta=0.25$) for the different retrieval techniques that we tried for bootstrapping our \textbf{Books-Small} corpus,}
\label{tab:retrieval-results}
\centering
\small
\begin{tabular}{l@{}rrr}
\toprule
Method & Precision & Recall & F$_\beta$-score \\
\midrule
Manatee & \textbf{100.00\%} & 17.34\% & \textbf{78.10\%} \\
Fuzzy Regexes & 78.98\% & 23.40\% & 69.30\% \\
Edit Distance & 74.00\% & \textbf{24.92\%} & 66.32\% \\
Concatenation & 72.50\% & 24.41\% & 64.97\% \\
BERTScore & 70.50\% & 23.74\% & 63.18\% \\
SentenceBERT & 69.50\% & 23.40\% & 62.28\% \\
Jaccard Similarity & 63.00\% & 21.21\% & 56.46\% \\
Reciprocal Rank Fusion & 62.00\% & 20.88\% & 55.56\% \\
Okapi BM25 & 35.03\% & 11.62\% & 31.31\% \\
\bottomrule
\end{tabular}
\end{table}

\noindent
\looseness=-1%  Make the paragraph one line shorter.
To select the best retrieval technique, we sampled 21 named entities from the abstracts and used each of the methods to produce up to 10 results. Then, three Czech experts employed as investigators in the \AHISTO{} project annotated the relevance of the results. Using the annotations, we computed Precision, Recall, and F$_\beta$-score ($\beta=0.25$), see Table~\ref{tab:retrieval-results}.

Based on F$_\beta$-score, we selected Manatee as the the best technique. The high accuracy of Manatee and Fuzzy Regexes shows that lemmatization and approximate search are important for the retrieval of named entities in OCR texts because they help with morphological variations and OCR errors.

\section{Annotator Instructions}
\label{app:annotator-instructions}

Figure~\ref{fig:test-set-annotation} shows the interface for the collection of manual annotations for the \textbf{Books-*} corpora.

Annotators were instructed to identify nested named entities, including territorial designations (e.g. Blažek of \emph{Kralupy}) and dedications of buildings (e.g. Church of \emph{St. Martin}). These nested annotations were utilized in the evaluation of NER models to prevent penalization for recognizing nested named entities as separate entities.

\begin{figure}[h]
\input figs/test-set-annotation
\caption{The interface for the annotation of our test dataset using the Google Documents web service. The interface includes annotator instructions (top) and an ordered list of sentences from books (bottom).}
\label{fig:test-set-annotation}
\vspace{-0.335cm}
\end{figure}

\begin{table*}[!t]
\caption{Example sentences in different languages from the \textbf{Abstracts-Tiny} and \textbf{Books-*} corpora. Manual annotations are compared to the predictions of \textbf{Model L}. Person names are \textbf{bold} and place names are \textit{in italics}. Nested named entities such as territorial designations and designations of buildings are \textbf{\textit{both bold and italic}}.}
\label{tab:example-sentences}
\footnotesize
\begin{tabularx}\linewidth{@{}llXX@{}}
\toprule
Corpus & Language & \multicolumn{1}{X}{Sentence} \\
&& Manually annotated ground truth & Model prediction \\
\midrule
Abstracts-Tiny & Czech &
\foreignlanguage{czech}{\textbf{Bohuněk} \textbf{a} \textbf{Kundrát,} \textbf{bratři} \textbf{z}~\textbf{\textit{Miroslavi,}} na základě svolení od \textbf{probošta} \textbf{\textit{dolnokounického}} \textbf{\textit{kláštera}} \textbf{Jana} a převorky, že mohou rozšířit rybník v \textit{Hlavaticích,} slibují jen na jistou vzdálenost od \textit{šumického dvora} zatopit a dovolit šumickým lidem, aby užívali tamní potok.}
&
\foreignlanguage{czech}{\textbf{Bohuněk} a \textbf{Kundrát,} bratři z \textit{Miroslavi,} na základě svolení od probošta dolnokounického kláštera \textbf{Jana} a převorky, že mohou rozšířit rybník v \textit{Hlavaticích,} slibují jen na jistou vzdálenost od šumického dvora zatopit a dovolit šumickým lidem, aby užívali tamní potok.}
\\
\midrule
Books-* & Czech &
\foreignlanguage{czech}{Vedle mnohaletého tohoto hejtmana čáslavského je tu \textbf{Žižkův} bratr \textbf{Jaroslav,} známí nám \textbf{bratří} \textbf{Valečovští,} sirotčí pozdější hejtmané \textbf{Jíra} \textbf{z}~\textbf{\textit{Řečice}} (\textit{Koudelova} u \textit{Čáslavě}) a \textbf{Blažek} \textbf{z}~\textbf{\textit{Kralup},} táborský \textbf{Jakub} \textbf{Kroměšín} a mnoho jiných statečných válečníků, i nejeden prostý voják, který však v~\textbf{Žižkově} radě zasedá jako rovný s~urozenými.}
&
\foreignlanguage{czech}{Vedle mnohaletého tohoto hejtmana čáslavského je tu \textbf{Žižkův} bratr \textbf{Jaroslav,} známí nám bratří \textbf{Valečovští,} sirotčí pozdější hejtmané \textbf{Jíra} \textbf{z} \textbf{Řečice} (\textit{Koudelova} u \textit{Čáslavě}) a \textbf{Blažek} \textbf{z}~\textbf{Kralup,} táborský \textbf{Jakub} \textbf{Kroměšín} a mnoho jiných statečných válečníků, i nejeden prostý voják, který však v~\textit{Žižkově} radě zasedá jako rovný s urozenými.}
\\
\midrule
Books-* & Latin &
\foreignlanguage{latin}{\textbf{Johannis} \textbf{Rupolth} vac., ad present. nobilis \textbf{Hinconis} \textbf{Berka} \textbf{de} \textbf{Duba} residentis in castro \textit{Scharffstein.} \textbf{Exec.} \textbf{pleb.} \textbf{in} \textbf{\textit{Arnorssdorff.}} C, IIII.- \textit{Horzielicz.-} Anno quo supra die XXVI April. data e. crida \textbf{Thome,} clerico de \textit{Antiqua} \textit{Boleslauia,} ad eccl. paroch.}
&
\foreignlanguage{latin}{\textbf{Johannis} \textbf{Rupolth} vac., ad present. nobilis \textbf{Hinconis} \textbf{Berka} \textbf{de} \textbf{Duba} \textbf{residentis} \textbf{in} \textbf{castro} \textbf{Scharffstein.} Exec. pleb. in \textit{Arnorssdorff.} C, IIII.- \textit{Horzielicz.-} Anno quo supra die XXVI April. data e. crida \textbf{Thome,} clerico de \textit{Antiqua} \textit{Boleslauia,} ad eccl. paroch.}
\\
\midrule
Books-* & German &
\foreignlanguage{german}{September 3. Der Rat zu \textit{Löbau} leiht 160 Schock zum Bau und zur Besserung der durch Brand und die Ketzer zerstörten Stadt. Nach Knothe Urkundenbuch von \textit{Kamenz} und \textit{Löbau} S. 253 (nach dem Original im \textit{Löbaner} \textit{Stadtarchiv,} jetzt im \textit{Hauptstaatsarchiv} zu \textit{Dresden).} 25 1432. September 12. Item \textbf{Nickel} \textbf{Windischs} ist ufgenomen des freitags vor des heiligen creucis exaltation.}
&
\foreignlanguage{german}{September 3. Der Rat zu \textit{Löbau} leiht 160 Schock zum Bau und zur Besserung der durch \textbf{Brand} und die Ketzer zerstörten Stadt. Nach Knothe Urkundenbuch von \textit{Kamenz} und \textit{Löbau} S. 253 (nach dem Original im \textit{Löbaner} \textit{Stadtarchiv,} jetzt im Hauptstaatsarchiv zu \textit{Dresden).} 25 1432. September 12. \textbf{Item} \textbf{Nickel} \textbf{Windischs} ist ufgenomen des freitags vor des heiligen creucis exaltation.}
\\
\bottomrule
\end{tabularx}
\end{table*}

\section{Training Details}
\label{app:training-details}

To fine-tune the XLM-RoBERTa-Base model, we used a learning rate of $5\cdot 10^{-5}$ with a linear decay until reaching 10 total training epochs or until convergence on the validation dataset. The fine-tuning took approximately 10 GPU hours on an NVIDIA Tesla T4 graphics card.

To fine-tune the XLM-RoBERTa-Large model, we used a smaller learning rate of $5\cdot 10^{-6}$ with a warm-up period of 20 epochs to mitigate overfitting and improve generalization performance. After the initial 20 epochs, we used a linear decay until reaching 200 total training epochs. The fine-tuning took approximately 74 GPU hours on an NVIDIA A40 graphics card.

% For both models, we used a batch size of 16 to fit into the available memory on the graphics card.

The total computational budget of our project was approximately 114 GPU hours.

\section{Model Predictions}
\label{app:model-predictions}

We randomly sampled three sentences from the testing split of the \textbf{Books-*} corpora, written in historical Czech, Latin, and German, as well as one sentence from the testing split of the \textbf{Abstracts-Tiny} corpus, written in modern Czech. Then, we compared the prediction of \textbf{Model L} with manual annotations for these four sentences.

Table~\ref{tab:example-sentences} illustrates that while \textbf{Model L} achieved a high level of precision, it failed to recall at least one named entity in every example sentence except the Czech sentence from the \textbf{Books-*} corpora. Furthermore, our model occasionally misidentifies the boundaries of named entities. An example of this can be observed in the Czech sentence from the \textbf{Books-*} corpora, where the named entity ``\foreignlanguage{czech}{bratří Valečovští}'' (translated as ``the Valečov brothers'') is incorrectly shortened to simply ``\foreignlanguage{czech}{Valečovští}'' (translated as ``the Valečovs''). We can see the opposite error in the Latin sentence, where the personal name ``\foreignlanguage{german}{Hinconis Berka de Duba}'' was incorrectly extended to include Berka's place of residence.

Compared to failures to recall named entities and identify their boundaries, errors in detecting types of named entities are rare as we already showed in Section~\ref{sec:qualitative-discussion}. The only example of an incorrectly predicted type occurs in the Czech sentence from the \textbf{Books-*} corpora, where the personal name ``\foreignlanguage{czech}{Žižkově}'' (translated as ``Žižka's'') was mistaken for Žižkov, a district of Prague, despite the presence of ample context clues in the surrounding sentence.

\end{document}